\documentclass[10pt,letterpaper]{article}
\usepackage{courier}
\usepackage{cogsci}
\usepackage{graphicx}
\usepackage[font=small]{caption}
\usepackage{amsmath}
\DeclareMathOperator{\argmax}{arg\,max} 

\cogscifinalcopy 

\usepackage{pslatex}
\usepackage{apacite}
\usepackage{float} 

\title{Flexible Compositional Learning of Structured Visual Concepts}
 
\author{{\large \bf Yanli Zhou (yanlizhou@nyu.edu)} \\
  Center for Data Science\\
  New York University
  \And {\large \bf Brenden M. Lake (brenden@nyu.edu)} \\
 Department of Psychology and Center for Data Science\\
  New York University}

\begin{document}
\setlength\titlebox{3cm} 
\maketitle

\begin{abstract}
Humans are highly efficient learners, with the ability to grasp the meaning of a new concept from just a few examples. Unlike popular computer vision systems, humans can flexibly leverage the compositional structure of the visual world, understanding new concepts as combinations of existing concepts. In the current paper, we study how people learn different types of visual compositions, using abstract visual forms with rich relational structure. We find that people can make meaningful compositional generalizations from just a few examples in a variety of scenarios, and we develop a Bayesian program induction model that provides a close fit to the behavioral data. Unlike past work examining special cases of compositionality, our work shows how a single computational approach can account for many distinct types of compositional generalization.

\textbf{Keywords:} 
concept learning; Bayesian inference; few-shot learning; visual learning; compositionality
\end{abstract}

\section{Introduction}
\begin{figure}[ht]
\centering
\includegraphics[width=\linewidth]{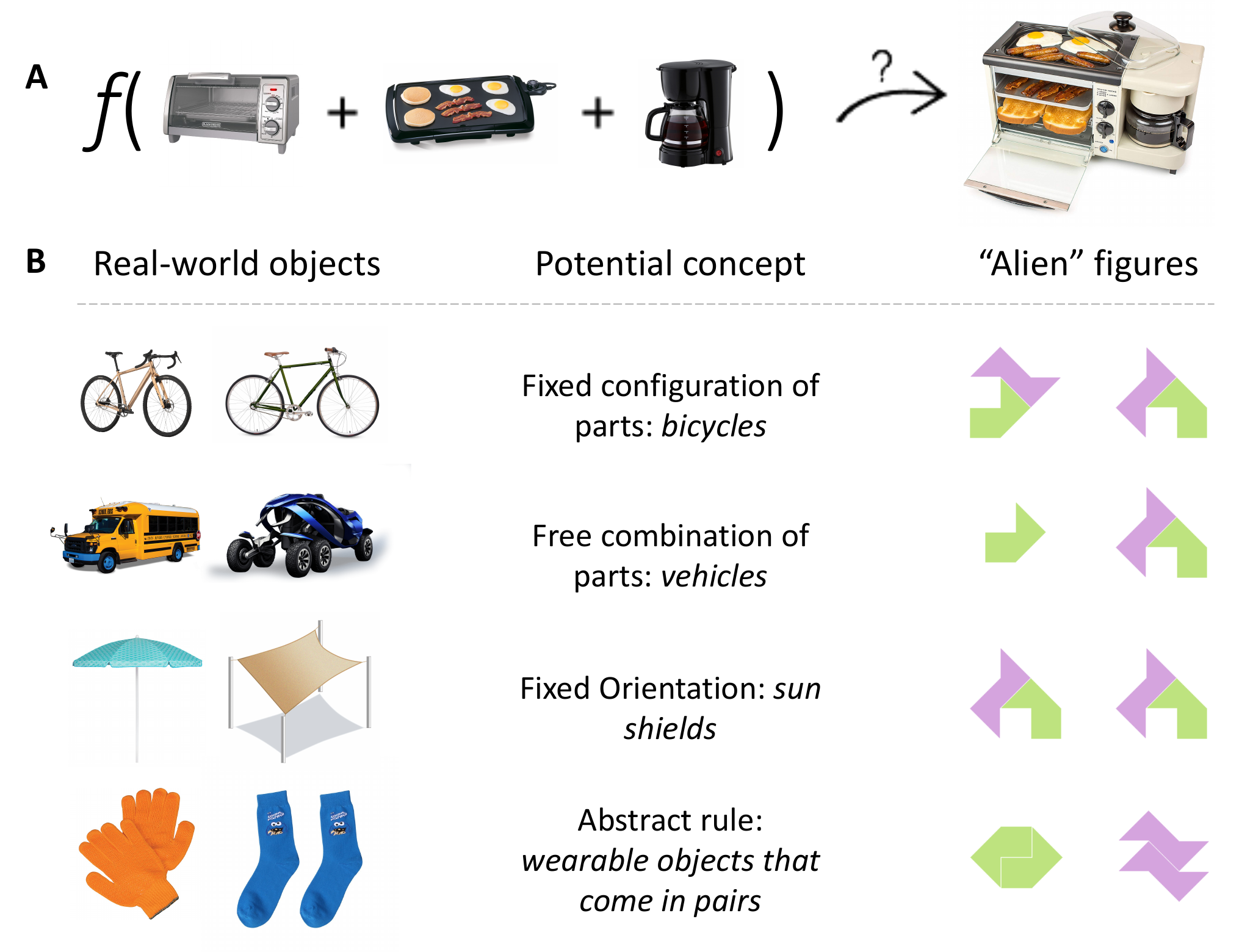}

\caption{Visual concept learning requires flexible notions of compositional structure. (A) Humans can learn the concept of \emph{breakfast machine} with a single example by recognizing familiar components and reasoning about their relations. Leading computer vision models tend to struggle with this concept. (B) Real-world visual concepts are defined by different types of compositions: 1. A \emph{bicycle} is a well-defined collection of parts in a consistent configuration; 2.  \emph{vehicles} allow a set of stereotyped parts to be combined more freely; 3. To be a \emph{sun shield}, an upright orientation is required; 4. \emph{Wearable objects that come in pairs} stipulate a repetition of \emph{wearable} elements. The rightmost column contains examples of experimental stimuli that are analogous to these concepts.}
\label{intro}
\end{figure}

Humans have a remarkable capacity to learn new concepts from limited data. Early in development, children can make meaningful generalizations from just one or few positive examples of a new word (\shortciteNP{smith_object_2002}; \citeNP{xu_word_2007}), an ability known as few-shot learning. Critical to few-shot learning is compositional generalization, the reuse and manipulation of preexisting knowledge of parts and relations to understand novel combinations (e.g., \shortciteNP{biederman_recognition-by-components:_1987}). For example, people who are familiar with \emph{coffee maker}, \emph{toaster oven} and \emph{griddle} can effortlessly grasp the concept of \emph{breakfast machine} upon seeing it for the first time (Fig. \ref{intro}A).\footnote{Example from Vicarious Research Blog.} On the other hand, computer vision models, while highly successful in many applications, are far more limited in their abilities to form compositional generalizations \shortcite{lake_building_2017}. For instance, a pre-trained ResNet-50 \shortcite{he_deep_2016} classifies the new concept in Fig. \ref{intro}A as a \emph{``waffle iron,''} whereas a strong image captioning system \cite{Xu2015show} describes it as \emph{``a close up of a toaster oven with some muffins in it."}

\begin{figure*}[ht]
  \includegraphics[width=\textwidth]{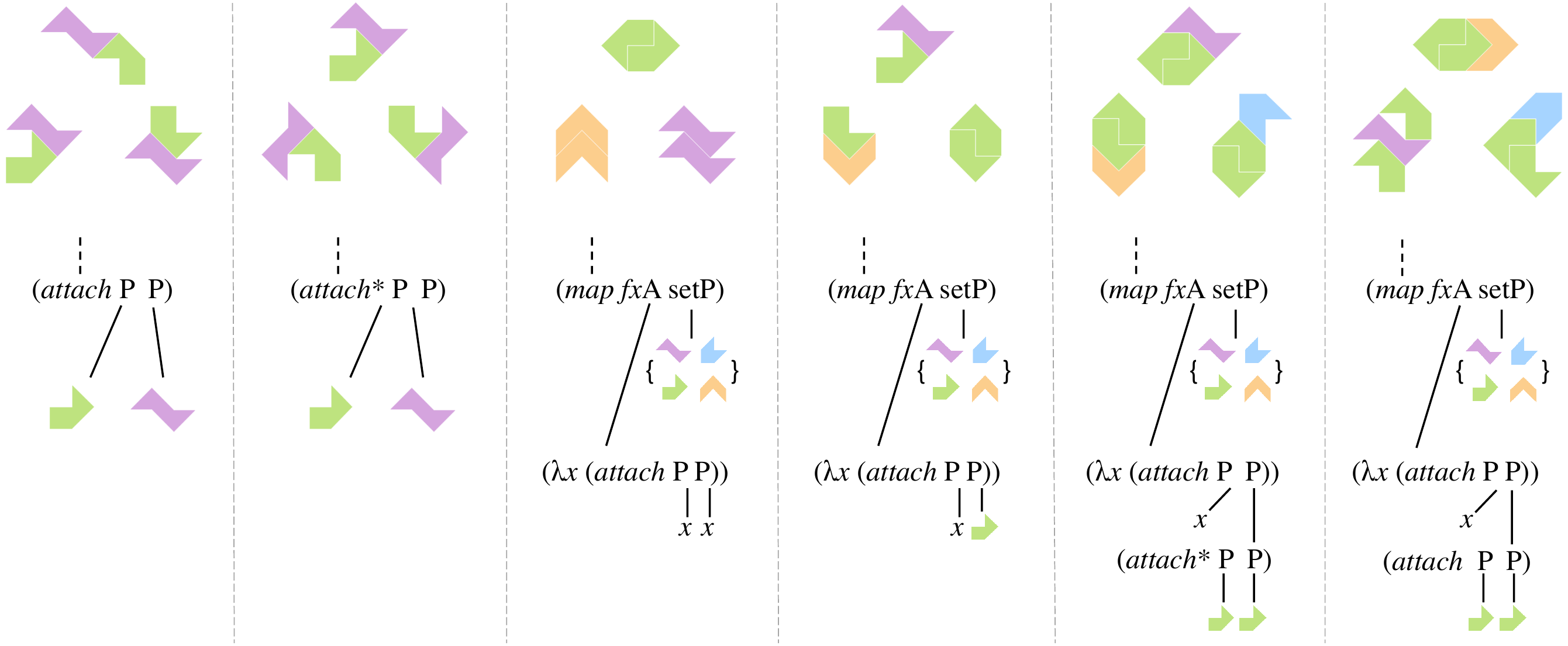}
  \caption{Examples of trial types tested in the experiments. Top: example alien figures given to participants to study on a given trial. Bottom: simplified parse tree of the most likely concept inferred by the Bayesian program induction model for each trial. The grammar over programs specifies primitive shapes and operations including the \emph{attach} function, which returns the set of all possible configurations of two parts, and the \emph{attach*} function which returns a set containing the single specified configuration. We can see that the spatial arrangement among components cannot be described with simple relations such as \emph{left}, \emph{right}, \emph{above} or \emph{under}.}
\label{concept}
\end{figure*}

There are qualitatively different types of composition present in real-world visual concepts, posing a challenging learning problem that demands manipulating parts and relations at various levels of abstraction (see examples in Fig. \ref{intro}B). A concept like \emph{bicycle} stipulates a fixed configuration of parts and relations (e.g.\ bikes have handlebars, a seat, and two wheels in a consistent configuration), whereas a concept like \emph{vehicle} allows category members to have freer combinations of parts and relations (varying numbers of wheels, motors, etc. are acceptable). A concept like \emph{sun shield} requires selectivity of object orientation, in order to fulfill a given conceptual constraint. Finally, a concept like \emph{clothing that comes in pairs} requires an additional degree of compositional abstraction, allowing a variety of parts to fill a role as long as they are duplicated.

Although previous work on few-shot learning has examined special cases of compositionality, we are still far from understanding the full variety of compositions present in real-world visual concepts (Fig. \ref{intro}B). In a seminal study, \citeA{xu_word_2007} examined word learning as Bayesian inference over tree-structured hypothesis spaces. Their model explains how children can make meaningful inferences from just a few examples, but compositional concepts were not considered.  \citeA{lake_human-level_2015} developed compositional models of learning handwritten characters, although individual characters are highly constrained in how their parts and configuration are allowed to vary (as in the 1\textsuperscript{st} row of Fig. \ref{intro}B). Other studies have considered sequential patterns \cite{overlan_learning_2017,Lake2019HumanFL} and recursive structures \cite{stuhlmuller_learning_2010,lake_recursive} more akin to the 4\textsuperscript{th} row of Fig. \ref{intro}B, or free combinations of parts akin to the 2\textsuperscript{nd} row of Fig. \ref{intro}B arranged in grid-like scenes \cite{orban2008}, although each of these concept types considered relatively simplistic spatial relations.

Our goal here is to study these various types of visual composition in a single experimental paradigm, and evaluate the success of Bayesian program induction in accounting for the inferences people make.
To do so, we consider a domain of visual concepts that is richly hierarchical, compositional, and relational. Using ``alien figures" as our stimuli, we conducted two experiments on few-shot concept learning, asking participants to make generalization judgements on test items after observing only a small number of positive examples. Following previous modeling work on Bayesian program induction in the visual domains \shortcite{stuhlmuller_learning_2010,lake_human-level_2015,overlan_learning_2017,lake_recursive}, we formalize learning in the alien categorization game as a search for the best programs for explaining the examples under a Bayesian score. The space of possible programs is constructed using a probabilistic language of thought (PLoT) \cite{goodman_rational_2008,piantadosi_2011,piantadosi_four_2016}, allowing for a wide range of compositions and abstractions. We found that our Bayesian program induction model provides an excellent account of experimental data, outperforming alternative models that lack key capacities to represent relations and compositionality. In addition, the fitted model parameters are psychologically meaningful, providing insight into people's inductive biases for these few-shot learning tasks.

\section{Behavioral Experiments}
Our experiments aimed to evaluate the flexibility of human compositional learning across a range of concept types. We adapted the few-shot learning paradigm of \citeauthor{xu_word_2007} \citeyear{xu_word_2007} for our purposes, as described below.

\subsubsection{Stimuli.} 
 The stimuli were described to participants as ``alien figures,'' which were programmatically generated by composing one to three shape primitives (see examples in Fig \ref{concept}). A composition of two parts is considered valid when they are non-overlapping and connected via two sides of identical length. Participants saw black-and-white outlines of each primitive\footnote{In this paper, all alien figures are shown with color-coded shape primitives for clarity.}. We left these primitives uncolored to motivate closer observations of stimulus shapes. As a visual aid in the experiment, rolling one's mouse over a primitive led all identical primitives in the display to become highlighted. The primitives were constructed through an additional level of compositionality, as they were composed of four isosceles right triangles. To form each set of training examples, we varied (1) which primitives can appear, (2) how many primitives appear in each exemplar (3) how the parts are composed and (4) if the configuration has a fixed orientation. 

\subsubsection{Task.} 
Participants took part in an ``alien figure categorization game" in which they are the assistant to a professor who collected samples of alien figures on a newly discovered planet. Their job in the game is to help the professor categorize a series of unnamed alien figures based on a small set of named examples. 

During each trial, participants were first familiarized with four different shape primitives. They were also informed that all relevant figures within the trial were built from these four primitives and no other primitives were possible. Next, participants were given a small set of example figures that shared a common name (see Fig. \ref{preds} for example trials). To minimize the effect of memory demands on learning, a display of the examples and primitives remained on screen throughout the trial. After an untimed observation period, participants entered a test stage in which they categorized a series of 9-13 unnamed alien figures. Specifically, participants chose `yes' or `no' for each test image to indicate whether it belongs to the same named category as the example images. We constructed each test set to cover a wide range of both possible and impossible extensions of potential concepts. 

We conducted two separate experiments with identical task procedures. The two experiments differed only in terms of the training and test sets in each trial. In Experiment 1, for every participant we tested 11 trials with each trial containing one to three training examples, followed by judgments on the test examples. Experiment 2 consisted of 10 trials and considered concept types that were more complex compared to those used in Experiment 1. We also used Experiment 2 to evaluate out-of-sample model predictions, since all model parameters were fit on the basis of Experiment 1. To study the effect of the exemplar set size on learning, participants in Experiment 2 were randomly separated into two conditions, based on whether they saw three or six exemplars of each concept. Trial orders were randomized for each participant; the set of allowable primitives were also randomized per participant per trial.

\subsubsection{Participants.} 

We used Amazon Mechanical Turk to recruit participants for both online experiments. Forty participants took part in Experiment 1, and 30 for each condition of Experiment 2. Responses from participants that failed one or more attention checks during either task were excluded. In the end, generalization judgements from 32, 25 and 20 participants were used in our reported analyses of Experiment 1, the 3-exemplar condition of Experiment 2, and the 6-exemplar condition of Experiment 2, respectively.  All participants finished the task within an hour and were paid \$5.00 at the completion of the experiment.

\section{Computational models}
We explored several types of computational models, with the aim of characterizing human generalizations in computational terms. 

\begin{figure}[t]
  \includegraphics[width=\linewidth]{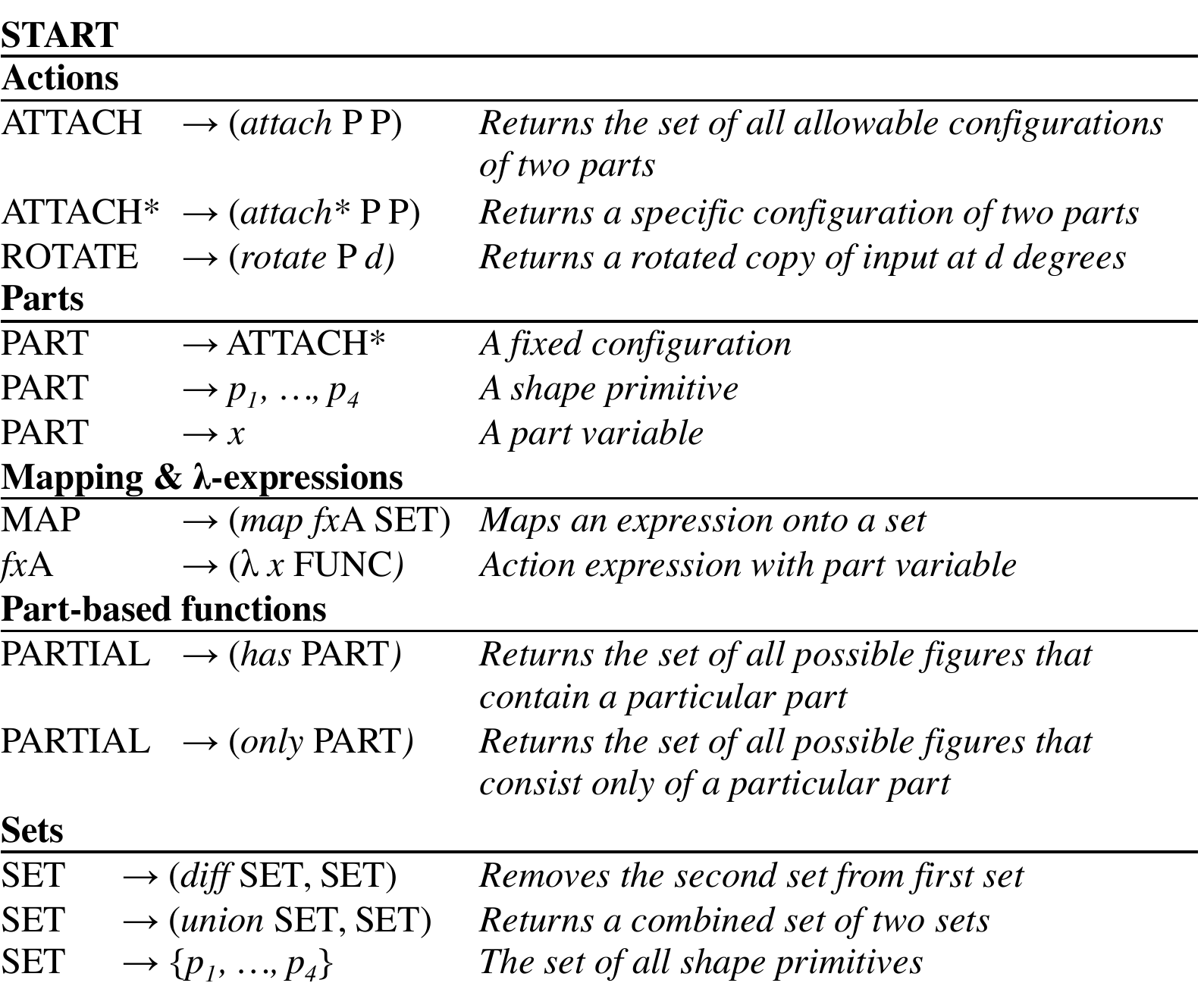}
  \caption{Core grammatical rules used to generate concept programs. The hypothesis space used in the study consisted of valid compositions of these primitives. Full grammar and supplementary material will be available online: https://github.com/yanlizhou/AlienFigures.}
\label{grammar}
\end{figure}

\subsection{Bayesian program induction}

To provide a unifying computational account of the wide range of generalization behavior elicited by various composition types, we developed a Bayesian program induction model that considers explicit, structural hypotheses as explanations for novel visual concepts. The model updates its beliefs over these hypotheses using a Bayesian framework \cite{goodman_rational_2008,piantadosi_four_2016} which generates human-like graded predictions with very limited data. In particular, the alien concepts were represented as probabilistic programs, which are structured generative models that produce distributions of examples. The goal of the learner is to infer programs consistent with the observed examples and the prior beliefs over programs. Inspired by  \shortciteA{piantadosi_2011} and \citeA{piantadosi_logical_2016}, we formed a compositional hypothesis space using a probabilistic grammar based on \emph{$\lambda$-calculus}. The grammar defines a set of primitive parts and operations which can be combined to build up programs of various levels of complexity (see Fig. \ref{concept} for examples of programs and output). Each sample from the grammar corresponds to a visual concept, and the production rules of the grammar specify the infinite space of possible concepts.

\subsubsection{Prior over programs.} To generate a concept, our grammar begins with expanding the START symbol into downstream nodes according to applicable rewrite rules. These nodes are subsequently rewritten until no further expansions are possible. Fig \ref{grammar} shows the core set of rules used to generate the programs (concepts) considered in our study. The output of each program is the set of all possible alien figures under such concept. In the example \texttt{(\emph{rotate} (\emph{attach\textsuperscript{1}} p\textsubscript{1} p\textsubscript{2}), 180)}, the inner most expression is first evaluated and returns the $1^{st}$ allowable configuration of primitives $p_1$ and $p_2$, which gets passed on to the outside expression that generates a rotated copy at $180^{\circ}$. This program has only a single element in its output set, as it corresponds to a generative process that fully specifies the types of parts, their configuration, and overall rotation. Figure orientation is based on four discrete possibilities, and two identical configurations at different rotations are considered distinct alien figures.  

The grammar also supports $\lambda$-expressions; together with mapping and set operations, the grammar can produce abstract concepts like \texttt{(\emph{map} (lambda x (\emph{attach} x x)) S)} which outputs the set of all possible configurations of two identical components sampled from the set $S$. Other function primitives in the grammar support hypotheses that do not fully specify a composition process. For example, \texttt{(\emph{has} p)} returns the set of all possible alien figures with $p$ as a part.

\begin{figure*}[t]
  \includegraphics[width=\textwidth]{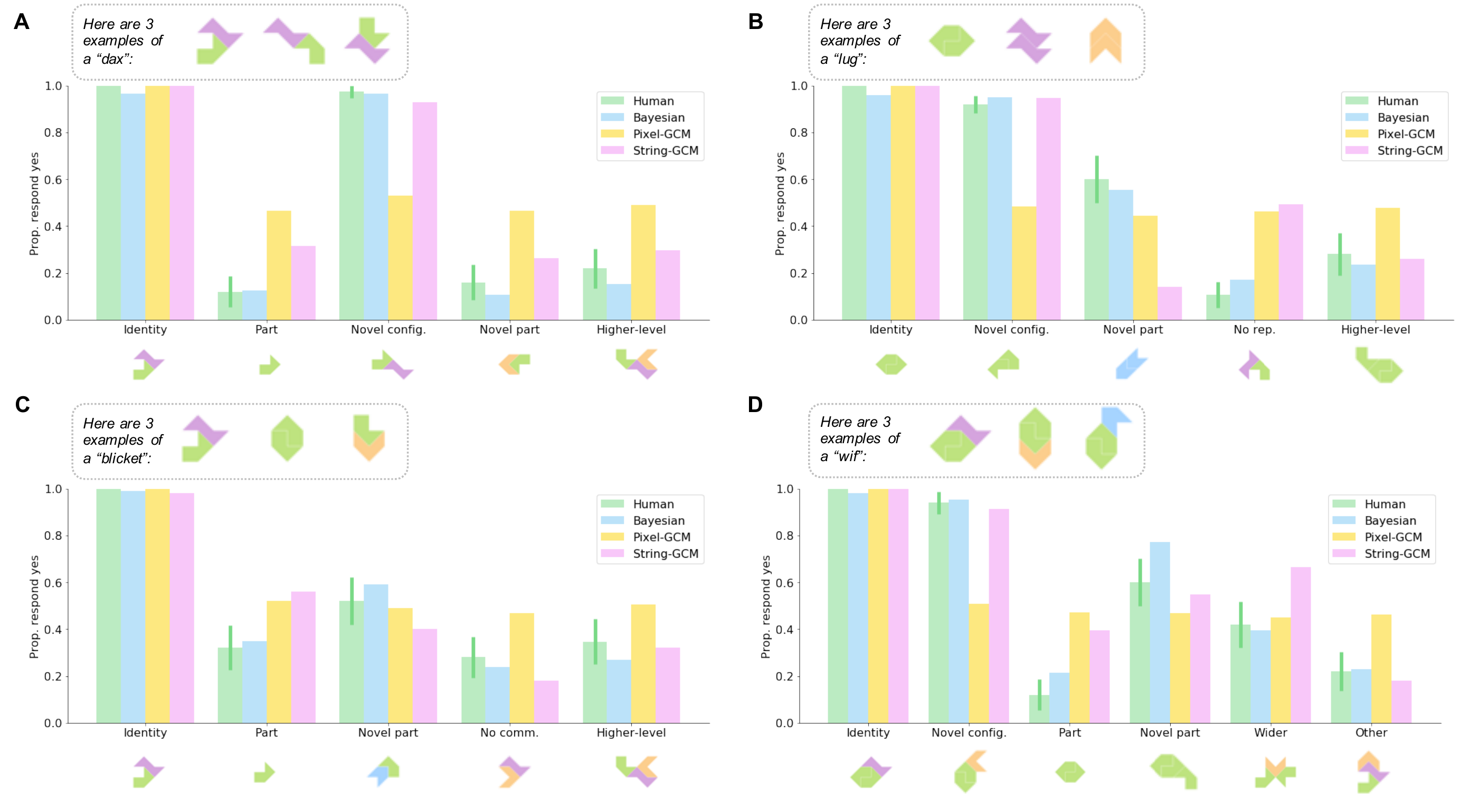}
  \caption{Model predictions on four of the trial types used in Experiment 2. The set of training examples is shown on the top of each panel; examples of test items were shown at the bottom. \emph{Identity} test items are identical to one of the examples; \emph{Part} test items are parts that appeared in one of examples; \emph{Novel configurations} items were new configurations of parts in examples; \emph{Novel part} items were conceptually consistent with examples but contained unseen parts; \emph{Higher-level} items were configurations with one of examples as subpart; \emph{No repetition}, \emph{No common} and \emph{Other} items in B,C and D were conceptually inconsistent with examples; \emph{Wider} items in D are samples from a wider concept for which the set of possible extensions is a superset of the concept of interest.} 
  
\label{preds}
\end{figure*}

\subsubsection{Likelihood and inference.} In Bayesian concept learning, the learner aims to compute the probability of a hypothesis $h$ given a set of examples $X=\{x_1,\dots,x_k\}$, or the posterior probability $P(h|X)$, which can be calculated by applying Bayes' rule: $P(h|X) \propto P(X|h)P(h)$. The first component, the likelihood of $X$ assuming hypothesis $h$ is true is defined as $$P(X|h) = \prod_{i}^{k}P(x_i|h) = \frac{1}{|h|^k},$$ where $|h|$ is the size of the concept. A likelihood function that is inversely proportional to the concept size, in our case the number of all unique outputs of a program, reflects the \emph{size principle} which assigns more weight to smaller hypotheses \cite{tenenbaum_bayesian_1999}. The second component $P(h)$, the prior probability of a concept, can be naturally derived from the grammar \cite{goodman_rational_2008}. Since each production of the grammar is a sequence of expansions of non-terminals, the probability of the production is the product of the probabilities associated with each expansion. This formulation operationalizes an important psychological preference for simplicity \cite{chater_simplicity:_2003} as shorter programs require fewer multiplications of expansion probabilities. To generate a model prediction for each test item $y$ after making a set of observations, we calculate the probability that the label $l_y \in \{0,1\}$ of $y$ is consistent with the set of observed examples $X$ as $$ P(l_y = 1 | X) = \sum_{h \in \mathcal{H}} P(l_y = 1 | h) P(h|X) $$ where $\mathcal{H}$ is the hypothesis space considered in our study. Approximate posterior inference was implemented in the LOTlib3 software package \cite{piantadosi2014lotlib}. For each trial, we ran three Monte Carlo chains for 100,000 steps of a tree-regeneration Markov chain Monte Carlo (MCMC) procedure \cite{goodman_rational_2008}.

\subsubsection{Parameter fitting.}

Given behavioral data collected in our experiments, we are interested in finding the set of grammar parameters that most likely generated people's generalization patterns. Formally, we would like to infer the probability of the set of parameters of interest, given human response data: $\argmax_{\vec{\theta},\alpha,\beta} P(\vec{\theta},\alpha,\beta | R, Y)$, where $\vec{\theta}$, $\alpha$ and $\beta$ are parameters of the learning model and $R$ is the set of human responses to the set of test items $Y$. To account for possible response noise in our collected generalization judgements, we fit a lapse rate $\alpha$, which determines the probability that a response was made at random. In the case of a lapse trial, we also represented a baseline preference for answering \emph{Yes} with parameter $\beta$. $\vec{\theta}$ is the set of grammar parameters, which are the probabilities associated with the distribution of expansions for each non-terminal. We only considered two grammar parameters that are psychologically meaningful, and we fixed the rest of expansions to have uniform probabilities. These two grammar parameters encode participants' preferences for \emph{orientation invariance} and \emph{configuration invariance}, respectively. We discuss the implications of the fitted values of these parameters in the Results section. The model-fitting procedure closely followed the one implemented by \citeA{piantadosi_logical_2016}, in which we performed stochastic search for the best fitting parameters via MCMC. The prior over the parameters used beta distributions with uninformative, uniform priors for $\vec{\theta}$, $\alpha$ and $\beta$.

\subsection{Alternative models}
We compare the Bayesian program induction model with two versions of an exemplar model known as the Generalized Context Model (GCM) \cite{nosofsky_attention_1986}. In a GCM, the probability of extending a category label $l_y$ to a new stimulus $y$ is based on its similarity to the training examples $X$:
$$P(l_y = 1 | X) \propto \frac{1}{k} \sum_{i}^{k}\exp(-w \cdot d(y, x_i))$$
where $d$ is a distance function and $w$ a scaling parameter. We evaluated two variants of the GCM with different distance measures.

\subsubsection{Pixel-GCM.} We used a deep convolutional neural net (CNN) to extract features of our visual stimuli from raw pixel data. A pre-trained 50-layer ResNet (He et al., 2015) was used to encode all images into vectorized representations. Cosine distance between two feature vectors was calculated as a measure of their similarity.

\subsubsection{String-GCM.} We also used a weighted Levenshtein distance to measure the distance between the string representations of two alien figures. For every image, its string format is a concatenation of 3 substrings that separately encode shape primitive types, primitive configurations and orientation. For example, an alien figure consisted of two primitives $p_1$ and $p_2$ connected according to their $1^{st}$ allowable configuration and rotated to $180^{\circ}$ can represented in the string format as ``$(p_1p_2)+1+180$''. We fit a weight parameter for each type of substring and the overall distance is a weighted average of the distances between each pair of corresponding substrings.

\begin{figure*}[ht]
\centering
\includegraphics[width=1.05\linewidth, trim={0mm 0mm 0mm 0mm}, clip]{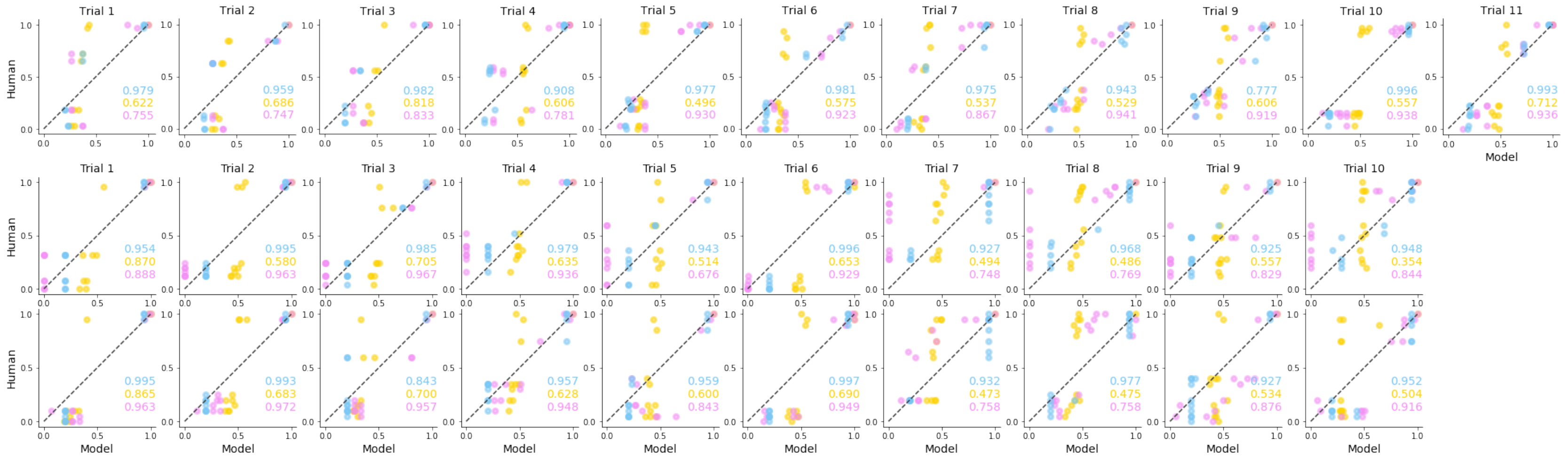}
\caption{Comparison between human responses and model predictions for each trial type of Experiment 1 (top row), 3-exemplar condition of Experiment 2 (middle row), and 6-exemplar condition of Experiment 2 (bottom row). Each dot in a scatter plot indicates the probability of responding `Yes' for each test item. The color of dots corresponds to the model type: blue for the Bayesian model, yellow for pixel-GCM and purple for string-GCM. Human-model correlations are also shown for each trial.}
\label{corr}
\end{figure*}

\section{Results}
The scatter plots in Fig. \ref{corr} summarize the correlations between human responses and model predictions for every trial type and model. Fig. \ref{preds} shows examples of model predictions alongside human data. Overall, we found that the Bayesian model provides an excellent account of human behavior with an average correlation of $r=0.955$ across all trial types studied in both experiments. We observe that the Bayesian model consistently assigns high probabilities to the test item that human participants found most likely, and produced graded predictions that tracked people's willingness to extend the concept. Importantly, the Bayesian model makes robust out-of-sample predictions. Using maximum-a-posteriori (MAP) parameters fit based on Experiment 1, the model was able to make good predictions for the more complex concepts in Experiment 2 ($r=0.952$ for Experiment 1, $r=0.948$ for 3-exemplar condition of Experiment 2 and $r=0.947$ for 6-exemplar condition of Experiment 2). All Bayesian model predictions regarding Experiment 2 trials reported were generated in this manner. 

\begin{table}
\begin{center} 
\caption{Fitted parameters values.} 
\label{table} 
\begin{tabular}{ll} 
\hline
Type    &  Probability \\
\hline
Orientation invariance    &   0.999 \\
Configuration invariance  &   0.725 \\
$\alpha$ (1-lapse rate) &   0.839 \\
$\beta$ (base rate for responding `yes')  &   0.714 \\
\hline
\end{tabular} 
\end{center} 
\end{table}

On the other hand, the two GCM variants fit the human data less closely, with average correlations of $r=0.604$ for the pixel-GCM and $r=0.874$ for the string-GCM. In the case of the pixel-GCM, the model responds strongly to the identity match, but unlike people, it does not clearly distinguish between the other types of generalization (Fig. \ref{preds}). The pre-trained CNN seemingly fails to perceive the stimuli in terms of their underlying parts and relations, at least without further fine-tuning. The string-GCM is a reasonably good account of the trial types with example figures sharing common parts, with no additional configuration constraints (e.g\ Fig. \ref{preds}A). This is unsurprising since the string format precisely encodes which shape primitives are present in each alien figure. The string-GCM struggles with more abstract rules that extends to unseen primitives (e.g.\ Fig. \ref{preds}B) or contain configurations of primitives not previously observed (e.g.\ Fig. \ref{preds}C). It also has a hard time grasping partial configuration constraints in a concept (e.g.\ Fig. \ref{preds}D).

The MAP values of fitted free parameters are reported in Table \ref{table}. Values of the two grammar parameters reveal two inductive biases people brought to bear when performing this visual concept learning task. The first parameter is the probability that a given concept contains images of the same configuration at different orientations. This probability is found to be very high, suggesting that our participants had a strong preference for orientation invariance when judging unnamed alien figures. People may have been influenced by their experience with named objects in the real world, which are usually orientation invariant. People were also biased towards concepts that do not require fixed configurations of parts. This is exemplified by their willingness to generalize to novel configurations, even when all examples shared the same configuration.

\section{Discussion}
We carried out an investigation of human few-shot visual concept learning, with an emphasis on concepts that compose primitives together in different ways. We studied ``alien figures'' that are richly structured, defined in terms of visual shapes connected in different systems of relations and at various levels of abstraction. Extending previous work on few-shot learning, we provided new empirical results on a set of concepts that better reflect the variety of ways parts combine in real wold visual concepts.

Our Bayesian program induction model provided predictions that closely matched human generalization patterns. Although the model is formulated exclusively to describe the class of alien figures, the model is flexible enough to be further extended by incorporating more or different primitives, or by adding grammatical rules to represent other types of visual concepts. Alternatively, we can formulate a set of different grammars and perform model comparison to distinguish between different language of thought theories within our existing framework.

Importantly, our paradigm is readily applicable to other learning approaches such as neural network (NN) models. The probabilistic grammar used in our studies can be used to sample many more concepts, as is needed for training NN models capable of few-shot learning through meta-learning \shortcite{vinyals2017matching}. By training NN models on this distribution of concepts, we can examine their ability to make compositional generalizations. We can also further refine NN models by fine-tuning them on a subset of the behavioral data, with the aim of better capturing more complex types of inductive bias. Direct comparisons between human and model behavior may further inform how to build machines with more compositional forms of learning \cite{Lake2019HumanFL}, and help identify potential ingredients that can endow NN models with more human-like capabilities.

In addition, various architectures and algorithms have been developed for problems such as Raven’s Progressive Matrices \shortcite{zhang_raven:_2019} and Bongard problems \cite{Nie2020Bongard}. In these datasets, simple visual forms are used to compose problems that test for compositional and relational reasoning abilities. Our task is related in some ways, but with a greater focus on understanding a variety of different types of composition. It's not obvious that models developed for these other domains will generalize to our tasks, but it's an important path for future work to consider.

Using our framework, we also plan to compare humans and computational models on generative tasks. Our Bayesian program induction model can generate new examples; however, generative tasks can provide a particularly direct window into human inductive biases, and it's likely that some modification will be needed to bring the prior closer to human expectations. We hope that generative tasks, building on the findings presented here, will further inform efforts to develop models of flexible, human-like compositional learning.

\section{Acknowledgements}
This work was supported by NSF Award 1922658 NRT-HDR: FUTURE Foundations, Translation, and Responsibility for Data Science. We thank Wai Keen Vong for helpful discussions of this manuscript. 

\bibliographystyle{apacite}
\setlength{\bibleftmargin}{.125in}
\setlength{\bibindent}{-\bibleftmargin}
\bibliography{references}


  
\end{document}